\documentclass[runningheads]{llncs}
\usepackage{graphicx}
\usepackage{dsfont}
\usepackage{algorithm}
\usepackage{algorithmic}
\usepackage{comment}
\usepackage{bbold}
\RequirePackage{subfigure}
\usepackage{amsmath,amssymb} 
\usepackage{tikz}
\usepackage{comment}
\usepackage{amsmath,amssymb} 

\usepackage{color}
\usepackage{hyperref}

\begin{document}
\pagestyle{headings}
\mainmatter
\def\ECCVSubNumber{173}  

\title{A Prescriptive Dirichlet Power Allocation Policy with Deep Reinforcement Learning} 

\titlerunning{A Prescriptive Dirichlet Power Allocation Policy with DRL}

\author{Yuan Tian \inst{1} \and
Minghao Han \inst{2} \and
Chetan Kulkarni \inst{3}\and
Olga Fink \inst{1}
}
\authorrunning{Y. Tian et al.}  
\institute{ETH Z\"urich\\
\email{\{yutian, ofink\}@ethz.ch}  \and
Harbin Institute of Technology \quad\quad\;\inst{3} NASA Ames Research Center \\
\email{mhhan@hit.edu.cn} \,
\email{chetan.s.kulkarni@nasa.gov}}

\maketitle
\begin{abstract}
Prescribing optimal operation based on the condition of the system and, thereby, potentially prolonging the remaining useful lifetime has a large potential for actively managing the availability, maintenance and costs of complex systems. Reinforcement learning (RL) algorithms are particularly suitable for this type of problems given their learning capabilities. A special case of a prescriptive operation is the power allocation task, which can be considered as a sequential allocation problem, where the action space is bounded by a simplex constraint. A general continuous action-space solution of such sequential allocation problems has still remained an open research question for RL algorithms. In continuous action-space, the standard Gaussian policy applied in reinforcement learning does not support simplex constraints, while the Gaussian-softmax policy introduces a bias during training. In this work, we propose the Dirichlet policy for continuous allocation tasks and analyze the bias and variance of its policy gradients. We demonstrate that the Dirichlet policy is bias-free and provides significantly faster convergence, better performance and better hyperparameters robustness over the Gaussian-softmax policy. Moreover, we demonstrate the applicability of the proposed algorithm on a prescriptive operation case, where we propose the Dirichlet power allocation policy and evaluate the performance on a case study of a set of multiple lithium-ion (Li-I) battery systems. The experimental results show the potential to prescribe optimal operation, improve the efficiency and sustainability of multi-power source systems.
\end{abstract}

\label{sec:introduction}
Prescribing  an optimal course of actions based on the current system state and, thereby, potentially prolonging the remaining useful lifetime has a large potential for actively managing the availability, maintenance and costs of complex systems \cite{ansari2019prima,ansari2020prescriptive,popp2020prescriptive}. In fact, it is a sequential decision making task that either requires very good dynamical models or models with very good learning capabilities. Reinforcement learning  (RL) algorithms have recently demonstrated a superior performance on sequential decision making tasks\cite{sutton1992reinforcement}. In particular, model-free RL, which estimates the optimal policy without relying on a model of the dynamics of the environment, has recently yielded very promising results in many challenging tasks ranging from gaming~\cite{mnih2013playing,silver2014deterministic}, to control problems~\cite{han2019h_,tian2022real}, prescriptive maintenance \cite{meissner2021developing} and AutoML~\cite{tian2020off}.

An important case of prescriptive operations for multi-power source systems is power allocation  with the goal of prolonging the lifetime or the usage time, thereby, improving the availability, maximizing the efficiency or minimizing the cost. These type of prescriptive operation tasks can be considered as sequential allocation problems. One of the major characteristics of allocation problems is that the action space is bounded by a simplex constraint. This constraint makes the application of RL algorithms in a continuous action space particularly challenging. Besides power allocation\cite{zhang2021integrated}, both sequential and single step allocation tasks are commonly encountered in several other application scenarios, such as task allocation\cite{deng2020task}, resource allocation \cite{feng2020joint,zhang2021resource}, order allocation \cite{feng2020integrated}, redundancy allocation\cite{zhang2021robust,nath2021evolutionary} and portfolio management \cite{jiang2017deep}. Particularly for allocation tasks involving complex systems, systems state and reliability considerations are crucial.


Several research studies have focused on allocation tasks with reinforcement learning \cite{xiong2018reinforcement,yang2018deep,jiang2017deep,maia2020regenerative}. However, one of the main limitations of the previously proposed RL approaches for allocation tasks is that they were solely able to operate in a discretized action space. This discretization typically precludes on the one hand fine-grained allocation actions since the number of discretized actions may become intractably high \cite{chou2017improving}. On the other hand, the action space needs to be carefully adjusted if the number of possible allocation options changes. These two aspects limit the scalability of the existing approaches substantially.

To enable a more general allocation decision-making, continuous action space  is required\cite{schulman2017proximal,haarnoja2018soft}. For continuous action space sequential allocation problems, the RL algorithms need to satisfy the simplex constraints as outlined above. However, the most commonly applied Gaussian policy in other RL tasks is not able to satisfy the simplex constraints\cite{lillicrap2015continuous,schulman2015trust,schulman2017proximal,haarnoja2018soft} . Gaussian-softmax policy could be an alternative solution \cite{jiang2017deep}. However, this function is not injective and has additional drawbacks such as its inability to model multi-modality \cite{joo2020dirichlet}. This leads to less efficient training and less effective performance.

In this paper, we focus on continuous action-space sequential allocation tasks and propose a Dirichlet-policy-based reinforcement learning framework for sequential allocation tasks. This enables us to overcome the aforementioned limitations. The proposed Dirichlet policy shows several advantages compared to the Gaussian, the Gaussian-softmax and the discretized policies. The Dirichlet policy inherently satisfies the simplex constraint. Moreover, it can be combined with all state-of-the-art stochastic policy RL algorithms. This makes it universally applicable for sequential allocation tasks. Ultimately, the proposed  Dirichlet policy exhibits good scalability and transferability properties. In this research, we theoretically demonstrate that the Dirichlet policy is bias-free and results in a lower variance in policy updates compared to the Gaussian-softmax policy. Finally, we experimentally demonstrate that the Dirichlet policy provides a significantly faster convergence, a better performance and is more robust to changes in hyperparameters compared to the Gaussian-softmax policy.

The performance of the proposed prescriptive operation framework in the context of sequential allocation problems is evaluated on a case study of multi-battery system applications with the goal of prolonging their working cycles. The developed framework only requires raw real-time current and voltage measurements, along with the incoming power demand as inputs. To the best of our knowledge, it is the first time an algorithm is capable of directly performing the load allocation strategy in an end-to-end way (without any involvement of model-based state estimation). We will demonstrate that compared to the equally distributed load allocation, the average length of the discharge cycle of the deployed four-battery system can be prolonged on average by 15.2$\%$ and of an eight-battery system by 31.9$\%$ over 5000 random initializations and random load profiles, making the batteries, thereby, more sustainable. Moreover, we will demonstrate that when implemented on degraded batteries in second-life applications with diverse degradation dynamics, the improvement becomes even more pronounced, reflecting a 151.0$\%$ extension of discharge cycles on average, enabling a reliable usage of second-life batteries. 

The contribution of this paper is two-fold: 1) We propose a novel RL-based solution for continuous action-space allocation tasks. In particular, we propose the Dirichlet policy, and demonstrate theoretically and experimentally its advantages. 2) Based on the proposed Dirichlet policy, we propose a prescriptive power allocation framework and evaluate its performance on multi-battery systems to prolong the service cycles of the power source systems. The developed framework shows the potential to improve the efficiency and sustainability of power systems with greater effectiveness.

\section{Related Work}
 Prescriptive operation is a comparably novel research direction that goes beyond just predicting the evolution of the system condition and the remaining useful life. The main goal of prescriptive operation is to develop algorithms that are not only able to predict the required measures but also to prescribe an optimal course of actions based on the current system state. Different objectives can be considered for prescriptive operation tasks, such as prolonging the remaining useful lifetime and, thereby, improving the reliability and availability of the system; completing a defined mission or reaching an operational goal also in case of adverse conditions or occurred faults; minimizing the emissions or minimizing the energy consumption. Several research studies have recently picked up the direction of prescriptive operation \cite{meissner2021developing,consilvio2019prescriptive}. For example, taking the economic and environmental impacts into account and prescribing maintenance operation to improve the efficiency of aircraft maintenance \cite{meissner2021developing}. For batteries, optimal charging schedules have been proposed to prolong the remaining useful life (RUL) \cite{sui2020multi}.  Prescriptive operation provides a very promising and urgently required research direction for the operation of industrial applications due to the increasing complexity and increasing requirements of complex industrial assets\cite{vater2019smart,ansari2019prima}. The prescriptive operation problems are in fact sequential decision making problems, for which RL methods have demonstrated very good learning capabilities \cite{meissner2021developing}.

In a reinforcement learning task, the agent observes the environment or system state and prescribes an action in order to maximize the cumulative expected future reward. The action space can be discrete, continuous, or mixed. The Q-Learning \cite{watkins1992q}, Deep Q-Network (DQN) \cite{mnih2015human} and its variants such as Double-DQN \cite{hasselt2010double} are normally designed for discrete action space tasks. To enable continuous action space, policy-based algorithms such as Proximal Policy Optimization (PPO) \cite{schulman2017proximal}, Trust Region Policy Optimization (TRPO) \cite{schulman2015trust} and Soft Actor-Critic \cite{haarnoja2018soft} have been proposed. These algorithms represent the stochastic policy by a Gaussian distribution, and the agent can sample from the distribution to get the specific action. Besides the stochastic policy, the Deep Deterministic Policy Gradient (DDPG) \cite{lillicrap2015continuous} uses a deterministic policy to tackle the continuous action space problem.  However, it shows a relatively weak performance in complex problems\cite{haarnoja2018soft}. Moreover, beta policy  has been proposed to improve the efficiency when physical constraints are present\cite{chou2017improving}. 

Allocation tasks are very commonly encountered in real-world prescriptive operation problems. Particularly the application of reinforcement learning to this type of problems and the elaboration of the theoretical perspective has remained relatively unexplored. The task is to find an optimal distribution of a limited resource given some defined goal and constraints. All allocation tasks need to fulfil the constraint that the action space is bounded by a simplex constraint. Examples of allocation tasks include computational resource allocation which is highly useful for emerging computational resource intense applications, such as industrial automation \cite{chen2020stackelberg}, blockchain applications \cite{feng2020joint}, UAV applications\cite{shimada2021novel}. Reliability redundancy allocation can help improving the system reliability and minimize the cost, weight or volume \cite{wang2020multi,sabri2019random}. Order allocation becomes more and more important to commercial companies like passenger transportation service companies \cite{kamandanipour2020stochastic,cao2020optimal}, food delivery companies \cite{sun2020research} and also other logistic companies \cite{jauhar2021proposed}. Optimal allocation directly influences the efficiency and the profit of such companies that are relying on limited resources. In the financial field, portfolio management is, in fact, also an allocation problem \cite{jiang2017deep}. 
Unfortunately, a general solution in RL for the allocation problems with the simplex constraint has still been missing and is an open research question.

An important application field of both allocation problems and prescriptive operation is power allocation \cite{hu2020battery} in multi-power source configurations which has recently been gaining in importance. A major challenge of power allocation strategies for multi-power source systems has been the design of optimal allocation strategies that take distinct observed states into account and consider different dynamics. For example in multi-battery systems, the individual batteries commonly start diverging in their states of health and remaining capacities \cite{zheng2015prognostics,severson2019data,hu2020battery} during the use. Small dissimilarities at the beginning of the lifetime may be amplified by different usage profiles. Once any of the individual batteries reaches the end of discharge (EoD), the normal operation of the entire system is impacted. Since individual batteries in the system may have dissimilar states of charge that are not directly measurable, distributing the power equally between all the batteries is not optimal. Allocating the power demand in an optimal way to each of the individual batteries has the potential to not only prolong the discharge cycle of the entire multi-battery system but also its lifetime and by that improve the sustainability of the batteries. 

Different types of power allocation strategies have been proposed including rule-based \cite{wang2019optimal,wang2019development,leonori2020optimization} and optimization-based approaches \cite{bai2019battery,zhang2016battery}. There are several limitations to these approaches. In the rule-based load allocation, each specific state would require the definition of customized rules. Thus, the rule-based approaches require extensive prior knowledge as well as extensive experiments for the different conditions that, for example, take the SoC or SoH into account, which cannot be measured directly. While they may be feasible for a small number of power sources in the system and a small range of possible operating conditions, it is difficult to scale them up to a larger number of power sources or to highly varying operating conditions and degradation dynamics. Since the feedback of such predefined rules is typically delayed, they are also hard to optimize, resulting in sub-optimal solutions.

Optimization-based approaches are vulnerable to uncertainties and changes in the schedule of the power profile, and to the best of our knowledge, they all rely on extracted features such as SoC. Furthermore, they require a relatively long online optimization, especially on high-dimensional allocation problems.

Machine learning approaches have also been increasingly applied to different battery management tasks, including predicting the future capacity \cite{nagulapati2021capacity,yang2018novel}, SoC \cite{liu2021battery,ng2020predicting,severson2019data,jiao2021more}, SoH, and remaining useful life(RUL) \cite{xu2021remaining}. In the power allocation domain, reinforcement learning-based approaches have also been recently investigated in a similar context \cite{xiong2018reinforcement}. Previous RL-based methodologies addressed the power allocation tasks by discretizing the action and state space, defining different weight combinations \cite{xiong2018reinforcement,xu2021novel,maia2020regenerative}. This significantly reduces their scalability and transferability ability. Due to the exploding action space problem, it is not feasible to directly increase the number of weight combinations for a more fine-grid decision-making \cite{bellman1956dynamic,lillicrap2015continuous}. Thus, to enable a more general power allocation strategy, continuous action space and corresponding approaches \cite{schulman2017proximal,haarnoja2018soft} are needed. 

\section{Methodology}
\label{sec:PM}
To solve the continuous action space allocation tasks, we introduce for the first time the Dirichlet policy. In the following, we first theoretically demonstrate the Dirichlet policy is bias-free with lower variance of policy updates than Gaussian-softmax policy. Moreover, we experimentally show that the Dirichlet policy provides a significantly faster convergence, a better performance and is more robust to changes in hyperparameters compared to the Gaussian-softmax policy. Moreover, we combine the Dirichlet distribution with the state-of-art soft actor-critic for the proposed Dirichlet Power Allocation Policy.

\subsection{Implications of the Gaussian Policy}

In reinforcement learning, a policy is always required to determine which action to take given the current state. In practice, the stochastic policy is usually paremetrized by a conditioned Gaussian distribution $\pi_\theta(\mathbf{x}|s) = \mathcal{N}(\mu_\theta(s), \delta_\theta(s))$, where $\mu$ and $\delta$ are outputs of the neural networks. However, the action $\mathbf{x}$ sampled from $\pi_\theta$ is not directly applicable to allocation tasks since the constraint $\sum_{i=0}^N a_i=1$ is not satisfied. It is straightforward to pass the generated candidate action $\mathbf{x}$ to a softmax function $\sigma$: $\mathbb{R}^N \rightarrow \mathbb{R}^N$ to obtain the allocation action:
\begin{equation}
    a_i =\sigma(x_i)_i =\frac{e^{x_i}}{\sum_{i=1}^N e^{x_i}}
\end{equation}

However, we show in the following that this approach would generate two side effects: the biased estimation and larger variance. Both of them would jeopardize the policy learning.

\subsubsection{Bias}
In allocation problems, the policy gradient is written as follows:
\begin{equation}
    \mathbb{E}g(\theta) =\mathbb{E}_s \int^{1}_{0} \pi(a|s)\nabla_\theta \log \pi(a|s)Q^\pi(s,a) \text{d}a
    \label{eq:true policy gradient}
\end{equation}

It should be noted that the softmax function is not injective, and many possible $\mathbf{x}$ can result in the same action $a$. More specifically, the softmax function is invariant under translation by the same value in each coordinate, i.e. $\sigma(\mathbf{x} +c\mathds{1}) = \sigma(\mathbf{x})$ for any constant $c \in \mathbb{R}$. When the softmax function is combined with the Gaussian policy to generate appropriate allocation actions, the distribution of $a$ is actually relevant to the distribution of the candidate action $\mathbf{x}$, and the probability density functions (PDF) satisfies
\begin{equation}
    \pi(a|s) = \int^\infty_{-\infty}\pi_\theta(\mathbf{x}+c\mathds{1}|s)\text{d}c 
\end{equation}

Substituting the above relation into the policy gradient follows that 
\begin{equation}
\begin{aligned}
    &\mathbb{E}g(\theta) =\mathbb{E}_s \int^\infty_{-\infty}\int^\infty_{-\infty}
    \\
    &\pi_\theta(\mathbf{x}+c\mathds{1}|s)\nabla_\theta\log\int^\infty_{-\infty}\pi_\theta(\mathbf{x}+c\mathds{1}|s)Q^\pi(s,\sigma(\mathbf{x})) \text{d}\mathbf{x}\text{d}c
\end{aligned}
\end{equation}
However, the policy gradient estimator $\mathbb{E}\hat{g}$ used in the ordinary RL algorithm is unaware of the inner integration over the scalar variable $c$, as in the following,
\begin{equation}
    \mathbb{E}\hat{g}(\theta) =\mathbb{E}_s \int^\infty_{-\infty}\pi_\theta(\mathbf{x}|s)\nabla_\theta \log \pi_\theta(\mathbf{x}|s)Q^\pi(s,\mathbf{x}) \text{d}\mathbf{x}
    \label{eq:policy gradient}
\end{equation}
As the mapping of the candidate action to the allocation action is done in the environment ( the specific allocation task),  the estimator is created based on the candidate action and inevitably introduces a bias. Even if assuming that the learned critic based on the candidate action can  predict the return precisely, i.e. $Q^\pi(s,\mathbf{x}) = Q^\pi(s,\sigma(\mathbf{x})),\forall x$, the bias still exists due to the unawareness of the marginalization over $c$.

One might also wonder whether using the transformed allocation action $a$ to compute the policy gradient can yield an unbiased estimation. Unfortunately, this is not the case. This would be equivalent to replacing the candidate action $x$ in \eqref{eq:policy gradient} with $a$. Though it looks similar to the form in \eqref{eq:true policy gradient}, the distributions $\pi_\theta$ and $\pi$ are not equivalent. In the end, this will only result in even more biased results.

\subsubsection{Variance}
In addition to the bias, the Gaussian policy also has the drawback that the variance of the policy gradient estimator is proportional to $1/\sigma^2$. This will induce the variance to reach infinity as the policy converges and becomes deterministic ($\sigma\rightarrow0$) \cite{chou2017improving}.

To illustrate this, a useful insight is gained by comparing the policy gradient with the natural policy gradient \cite{kakade2001natural}. The policy gradient in \eqref{eq:policy gradient} does not necessarily produce the steepest policy updates \cite{amari1998natural}, while the natural policy gradient does. The natural policy gradient is given by 
\begin{equation}
    g_{\text{nat}}(\theta) = \mathbb{E}_s F^{-1}(\theta)\hat{g}(\theta)
\end{equation}
where $F$ denotes the Fisher information matrix, defined as 
\begin{equation}
    F = \mathbb{E}_{a\sim\pi_\theta}\left[\nabla_\theta \log \pi_\theta(a|s) \nabla_\theta \log \pi_\theta(a|s)^T\right]
\end{equation}
The policy gradient vector is composed of the length and the direction. The ordinary policy gradient may have the correct direction but not necessarily the correct length. The natural policy gradient adjusts the learning rate according to the policy distribution and produces the steepest step.
As shown in \cite{chou2017improving}, the Fisher information matrix for Gaussian policy is proportional to $1/\sigma^{2}$, which implies that the more deterministic the policy is, the smaller the update step that should be taken. In the end, the constant update steps will overshoot and increase the variance of the policy gradient estimator.
\subsection{Dirichlet Policy}

Since the general Gaussian or Gaussian softmax are not directly applicable to the optimization of allocation problems, directly applying standard reinforcement learning frameworks or other control algorithms to allocation tasks will result in sub-optimal results that suffer from excessive parameter tuning and/or model complexity. To improve the stability and convergence speed of optimization tasks of allocation problems in continuous action spaces, we propose parameterizing the policy using Dirichlet distribution, which inherently satisfies the simplex constraint and enables an efficient optimization of allocation tasks in continuous action spaces:

\begin{equation}
    \pi_{\theta}(a|s) = \frac{1}{B(\alpha)}\prod_{i=1}^K a_{i}^{\alpha_i-1}
\end{equation}
where $B(\alpha)$ denotes the multivariate beta function and can be expressed in terms of the gamma function $\Gamma$ as follows:
\begin{equation}
    B(\alpha)=\frac{\prod_{i=1}^K \Gamma(\alpha_i)}{\Gamma(\sum_{i=1}^K\alpha_i)}.
\end{equation}
Here, the distribution is shaped by the shape parameters $\alpha$, which is the output of the neural network $f_\theta(s)$. Thus, the policy is eventually determined by $\theta$.

\subsubsection{Variance of the Dirichlet Policy} 
Let $A=\sum_{i=1}^K\alpha_i$
\begin{equation}
     \log \pi_\theta(a|s)=\log(\Gamma(A))-\sum_{i=1}\log(\Gamma(\alpha_i)) + \sum_{i=1}(\alpha_i-1)\log(\Gamma(a_i))
\end{equation}
Taking the fact that $\partial A/\partial \alpha_i =1$ and $\partial \alpha_j/\partial \alpha_i =0$ into account results in:
\begin{equation}
    \frac{\partial\log \pi_\theta(a|s)}{\partial\alpha_i} = \psi(A) - \psi(\alpha_i) + \log(a_i)
\end{equation}
with the second order derivative 
\begin{equation}
    \frac{\partial^2\log \pi_\theta(a|s)}{\partial\alpha_i\partial\alpha_j} = \psi'(A)- \psi'(\alpha_i)\delta_{ij}
\end{equation}
where $\psi'(z)=\psi^{(1)}(z)$ and $\psi^{(m)}(z) = \frac{d^{m+1} }{dz^{m+1}}\ln \Gamma(z)$ is the polygamma function, the $m_{\text{th}}$ derivative of the logarithm of the gamma function.

According to the regularity conditions \cite{wasserman2013all}, the Fisher information matrix can also be obtained from the second-order partial derivatives of the log-likelihood function,
\begin{equation}
\begin{aligned}
    F(\alpha) =& -\mathbb{E}_{a~\pi_\theta} 
    \begin{bmatrix}
        \frac{\partial^2\log \pi_\theta(a|s)}{\partial\alpha_1\partial\alpha_1}  &
        \cdots 
        &
        \frac{\partial^2\log \pi_\theta(a|s)}{\partial\alpha_1\partial\alpha_K} 
        \\
        \vdots 
        & 
        \ddots
        &
        \vdots
        \\
        \frac{\partial^2\log \pi_\theta(a|s)}{\partial\alpha_K\partial\alpha_1}
        &
        \cdots
        &
        \frac{\partial^2\log \pi_\theta(a|s)}{\partial\alpha_K\partial\alpha_K} 
    \end{bmatrix}\\
   = &\begin{bmatrix}
        \psi'(\alpha_1) - \psi'(A)
        &
        \cdots 
        &
         - \psi'(A)
        \\
        \vdots 
        & 
        \ddots
        &
        \vdots
        \\
         - \psi'(A)
        &
        \cdots
        &
        \psi'(\alpha_K) - \psi'(A)
    \end{bmatrix}\\
    \end{aligned}
\end{equation}
The variance of the Dirichlet policy is given by $Var[a_i] = \frac{\alpha_i(A-\alpha_i)}{A(A+1)}$.
As the policy becomes deterministic given different states, certain allocation actions $\alpha_i$ and $A$ approach infinity simultaneously. As shown in  \cite{chou2017improving}, $\psi'(z)$ goes to zero as $z$ goes to infinity. Thus, the inverse of the Fisher information matrix goes to infinity. This ensures that the update steps will not overshoot and the variance of the policy gradient goes to zero.

To summarize, the Dirichlet policy can intrinsically produce unbiased policy gradient estimations, while the variance of policy updates is also guaranteed to be lower than that of the Gaussian policy. These are both favorable properties to enhance the convergence speed and allocation performance.
\subsection{Simplex regression experiment}
\begin{figure}[htbp]
\centering
\includegraphics[width=0.5\columnwidth]{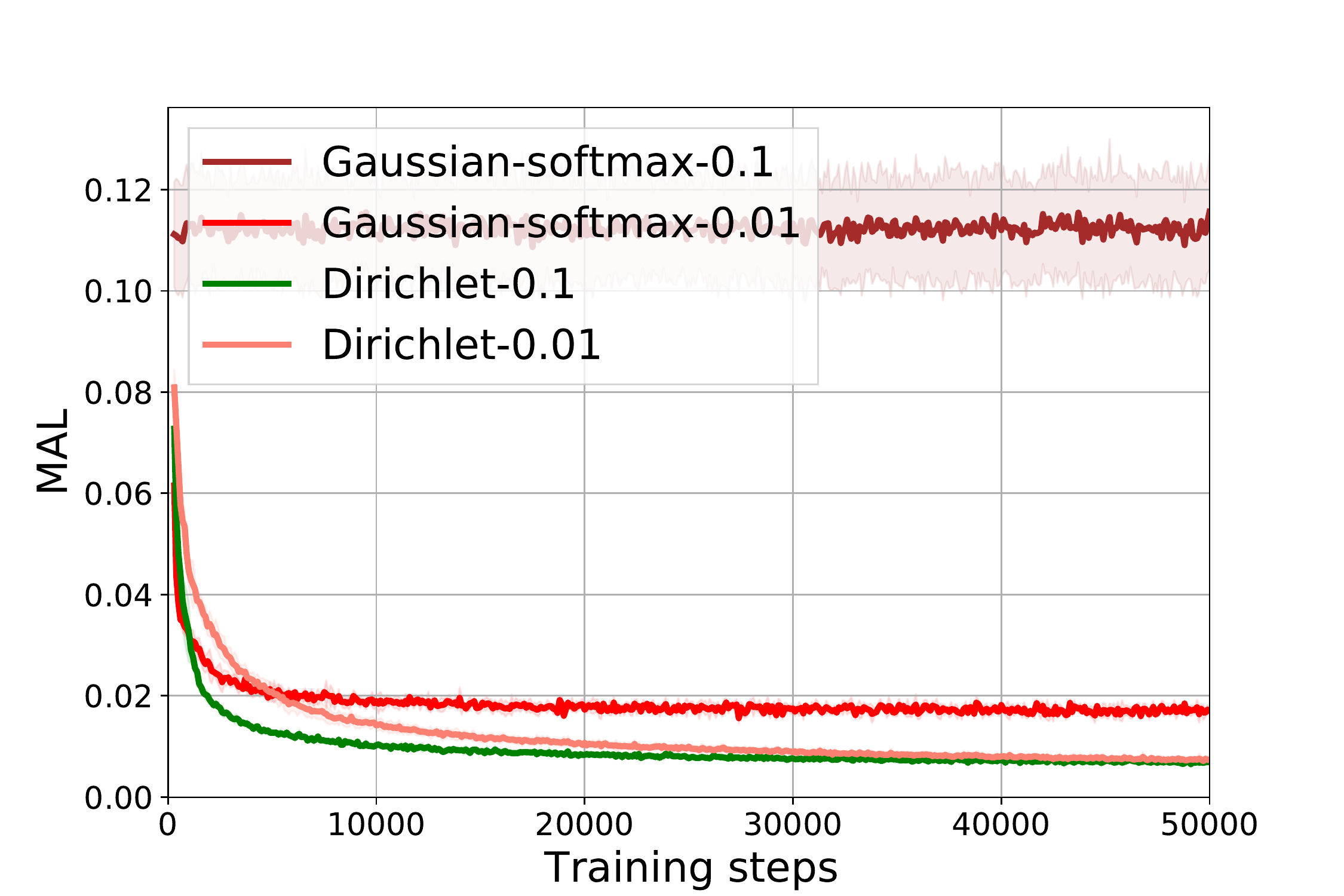}
\caption{{\textbf{Numerical experiment results} We compare the learning curves of both output layers with two different learning rates: $0.1$ and $0.01$.}}
\label{fig:NE}
\end{figure}

To demonstrate the efficiency and effectiveness of the proposed methodology, we evaluate it first on a simple simplex regression task. The objective is to reconstruct and sequence a 4-dimension simplex from a 3-dimension vector where a dimension is randomly removed from the target 4-dimension simplex. For example, given a random simplex vector $[0.4,0.2,0.3,0.1]$, after a dimension is randomly removed, the input data becomes $[0.4,0.3,0.1]$. The target output is then the ranked reconstructed  simplex $[0.1,0.2,0.3,0.4]$. We use the Mean Average Error (MAE). We apply on the hand the proposed Dirichlet policy framework and compare it to the Gaussian-softmax policy. The result shows that the Dirichlet distribution performs better and is more robust to hyperparamters. The Dirichlet policy performs two times better compared to Gaussian-softmax policy with a learning rate of $0.01$. Besides, the Dirichlet policy is more robust against different learning rates, while the Gaussian-softmax policy failed with a high learning rate $0.1$. see figure \ref{fig:NE}.

For the neural networks in the numerical experiment, we use a fully connected multi-layer perceptron (MLP) with two hidden layers of 64 units each, outputting the $\alpha$ of a Dirichlet distribution or the $\mu$ and $\sigma$ of the Gaussian distribution. For the Dirichlet distribution network, the first layer uses leaky-ReLU as the activation function. The second layer uses tanh as the activation function. The $\alpha$ is modeled by a softplus element-wise operation with $\log(1 + \exp (x))$. A constant 1 is added to the output to make sure that $\alpha \geqq 1$ . The choice of the activation function is motivated by the design of the Beta policy \cite{chou2017improving}. For the Gaussian-softmax network, the hidden layers have leaky-ReLU as the activation function, while the final output layer is mapping to a simplex with a softmax function. 

\subsection{Soft Actor-Critic}
In this paper, we applied the off-policy reinforcement learning algorithm soft actor-critic (SAC) \cite{haarnoja2018soft} with the proposed Dirichlet policy. The SAC is based on the maximum entropy reinforcement learning framework\cite{haarnoja2017reinforcement}, where the objective is to maximize both the entropy of the policy and the cumulative return. As a result, it significantly increases training stability  and improves exploration during training. Furthermore, it was demonstrated to be 10 to 100 \cite{haarnoja2018soft} times more data-efficient compared to any other on-policy algorithms applied to traditional RL tasks. 

For the learning of the critic, the objective function is defined as:

\begin{equation}
    J(Q) = \mathbb{E}_{(s,a)\sim \mathcal{D}}\left[\frac{1}{2}(Q(s,a)-Q_{target}(s,a))^2\right]
\end{equation}
where $Q_{target}$ is the approximated target of $Q$ :

\begin{equation}
\begin{aligned}
&Q_{target}(s,a) = R(s,a) +\\
&\gamma [Q_{target}(s', f(\epsilon,s'))-\beta \log \pi(a'|s')]
\end{aligned}
\end{equation}
The objective function of the the policy network is given by:

\begin{equation}
\begin{aligned}
J(\pi) = \mathbb{E}_{ \mathcal{D}}\left[ \beta [\log(\pi_\theta(f_\theta(\epsilon,s)|s))]-Q(s,f_\theta(\epsilon,s)) \right]
\label{SAC}
\end{aligned}
\end{equation}
where $\pi_\theta$ is parameterized by a neural network $f_\theta$, $\epsilon$ is an input vector, and the $\mathcal{D}\doteq\{(s,a,s',r)\}$ is the replay buffer  for storing the MDP tuples~\cite{mnih2015human} and $\beta$ is a positive Lagrange multiplier that controls the relative importance of the policy entropy versus the cumulative return. 
\subsection{Hyperparameters setting}
For the following experiments, we combine the proposed Dirichlet policy with the SAC framework. For the policy network, we use the same architecture design as for the toy experiment with the difference that 256 units are used. And for the Q-network, we use a fully connected MLP with two hidden layers of 256 units, outputting the Q-value. All the hidden layers use leaky-ReLU as the activation function.

In our implementation, the double Q-learning technique \cite{van2016deep}
is exploited, whereby two Q-functions $\{Q_1,Q_2\}$ are parameterized by neural networks with parameters $\nu_1$ $\nu_2$. The Q-function with the lower value is exploited in the policy learning step \cite{fujimoto2018addressing}, which is useful in mitigating performance degradation caused by the bias in the value estimation. 

The optimization of the networks' weights was carried out with the \textit{Adam} algorithm. The \textit{Kaiming} initializer was used for the weight initializations \cite{he2015delving}. Table \ref{tb:settings_SAC} provides a detailed overview of the hyperparameters used for the experiments. Training is conducted on a 2.3 GHz 8-core Intel Core i9 CPU.

\begin{table}[ht]
\begin{center}
\caption{SAC Hyperparameters}
\begin{tabular}{l|c}
\hline
Hyperparameters                   & Value   \\ \hline
Minibatch size                    & 1024     \\\hline
Learning rate - Actor             & 1e-4    \\\hline
Learning rate - Critic            & 3e-4    \\\hline
Target entropy                    & -$\sqrt{d}$ \\\hline
Target smoothing coefficient($\tau$) & 0.005\\\hline
Discount($\gamma$)                & 0.99\\ \hline
Updates per step & 1
\label{tb:settings_SAC}
\end{tabular}
\end{center}
\end{table}

\section{Power Allocation Case Study}
\label{sec:Experiments}
To further evaluate the performance of the proposed method, we design a case study of multi-battery system applications with the goal to prolong their working cycles. We assume that the power allocation can be controlled at the level of a single battery and that no cell-balancing is applied. 

The information most commonly used in battery health-related analytics is the operating current and voltage measurements collected by standard battery management systems \cite{richardson2018gaussian,severson2019data}. In this case study, we aim to utilize only raw measurements of current and voltage directly measured on the batteries (before the DC-DC converter) and extend the capability of machine learning from descriptive and predictive analytics to end-to-end prescriptive decision-making. To the best of our knowledge, it is the first time an algorithm is capable of directly performing the load allocation
strategy in an end-to-end way (without any involvement of model-based state estimation)

The objective of the desired power allocation strategy is to prolong the working cycle of the deployed multi-battery system. To achieve this, we formulate this problem as a Markov decision process (MDP) and propose to solve it with the Dirichlet policy reinforcement learning. 

Every operation or maneuver of a multi-battery device will impose a power demand $P_t$ on the system. The RL-based strategy will prescribe an action $a_t$ that dynamically allocates the power demand $P_t$ based on the observed state $s_t$. In our case, $s_t$ is represented by the real-time operational current and voltage of all the batteries in the system and the total power demand, resulting in $s_t = [V_t, I_t,P_t]$. Then, the system's state changes according to the allocation strategy and the system dynamics $\mathcal{P}$. To achieve the objective of prolonging the working cycle of the battery device, we provide a reward of $r_t=1$ to the agent at each time step at which all the batteries in the system are still operational or the voltages are all higher than the end-of-discharge (EoD) state. Given a discount factor $\gamma\leq 1$, an optimal allocation strategy maximizes the expected discounted sum of future rewards, or return:
\begin{equation}
  R_{\tau} = \mathbb{E}[\sum_{t=0}^{\inf}\gamma^tr(s_t,a_t)|s_0=s]
\end{equation}
where $\mathbb{E}$ indicates the expected value. $R_s$ characterizes the long-term value of the allocation strategy from an initial state $s_0$ onwards.

\begin{figure}[htbp]
\centering
\includegraphics[width=\columnwidth]{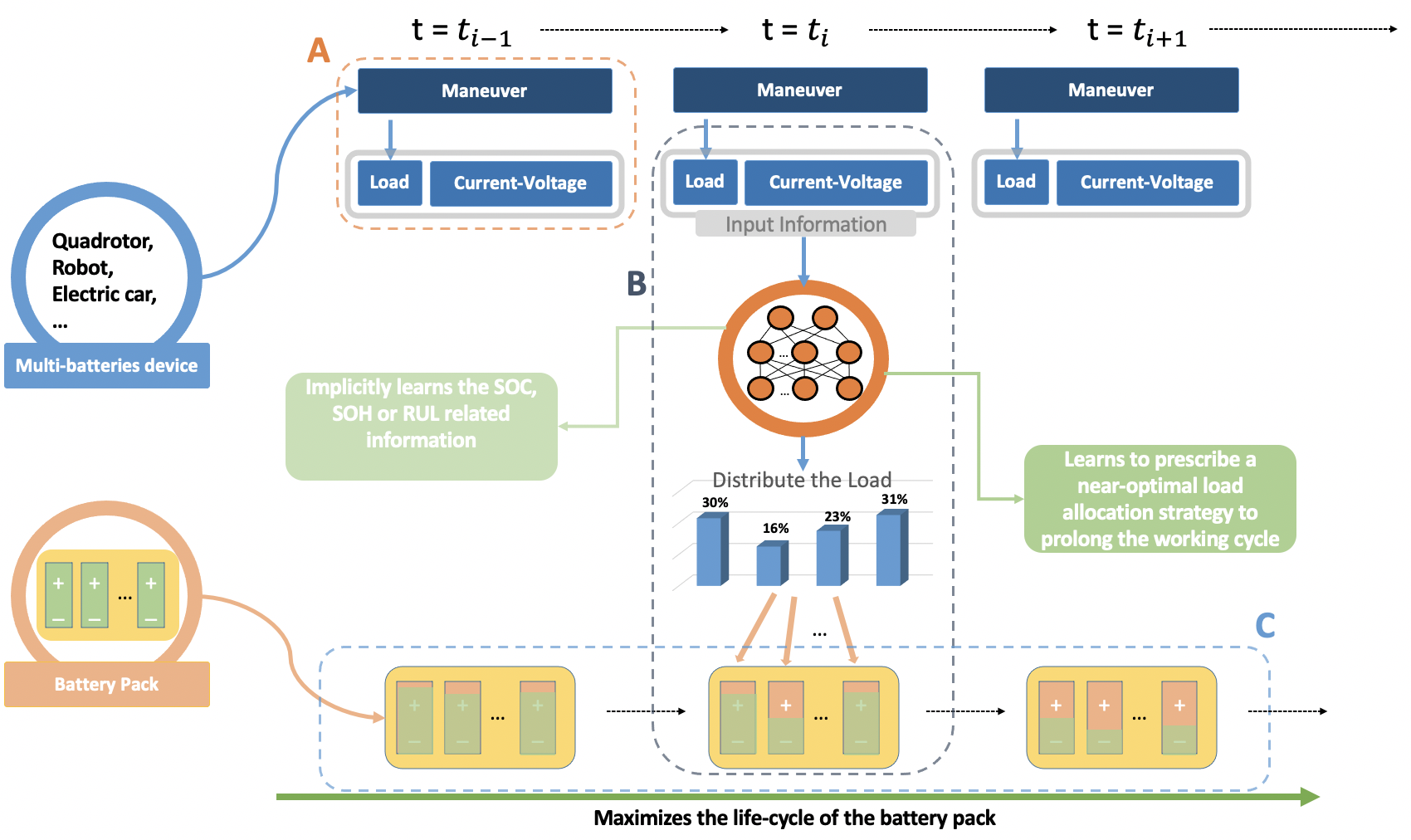}
\caption{{\textbf{Overview of the power allocation for multi-battery systems.}}}
\label{fig:MainFIG}
\end{figure}

Figure \ref{fig:MainFIG} is the overview of the Dirichlet power allocation framework. \textbf{A:} When deploying the proposed strategy on any device with a multi-battery system, such as a quadrotor, a robot, or an electric car, any maneuver induces a load demand. For every maneuver, the trained strategy receives the incoming load demand with the real-time current-voltage measurement. It distributes the power only based on the received information or observation without any online optimization. \textbf{B:} The proposed strategy is represented by a neural network, which takes the measurements as input and outputs a weight combination on how the load should be distributed to the individual batteries. The trained network can dynamically allocate the power in an end-to-end way without any estimation of the degradation state. With the input information of current and voltage measurements, it can first implicitly learn the health of the batteries,  such as SOC, SOH, or RUL, for decision-making. With the proposed Dirichlet policy, which inherently satisfies the \textbb{simplex} constraint of the allocation tasks, it can prescribe fine-grid allocation weights in a continuous manner and can be trained more efficiently and effectively. \textbf{C:} In this paper, the objective is prolonging the working cycle of the deployed multi-battery systems, which could be changed in other tasks according to different requirement.

\subsection{Simulation Environment}
We train the allocation strategy in a simulation environment. The simulation environment is a multiple Li-I battery system computational model from the NASA prognostic model library \cite{daigle2013electrochemistry,nasaBattery}. For an individual battery, the state changes over time as a function of input load and current system states. The load is allocated at the level of a single battery cell and no load balancing is performed. This computational model serves as a reliable proxy for actual battery dynamics and allows for fast iterations over the controller design. It is worth mentioning that batteries generally have a relatively complex working dynamics, which is also a challenging case study by which to evaluate the general performance of a power allocation strategy. 

For training, we randomly sample battery states during operation as initial states $s_0$ for any new episode.  The episode will be re-initialized when any of the batteries reaches the EoD state. 

\subsection{Results}

The proposed framework is evaluated with respect to three performance aspects: 1) Performance is assessed on a case study comprising a four Li-I batteries system.  2) Scalability is evaluated on an eight Li-I batteries system. 3) Transferablity is evaluated on a four second-life Li-I batteries system where each of the batteries exhibits different degradation dynamics. 4) We compare the learning performance to other state-of-the-art reinforcement learning methods, 5) and also to heuristic strategies. All the performance metrics are averaged among 5000 different random initializations with random load profiles, see Table \ref{tb:main}.
\begin{table}[ht]
\begin{center}
\begin{tabular}{l|c}
\hline
Experiments                    & Average Improvement  \\ \hline
four-battery system       & 15.2$\%$   \\ \hline
eight-battery system             & 31.9$\%$\\  \hline
four-second-life-battery system             & 151.0$\%$

\label{tb:main}
\end{tabular}
\end{center}
\caption{Main results of the average improvement between the baseline among 5000 different random initialization with random load profiles}
\end{table}

\begin{figure*}[ht]
    
    \centering
    \includegraphics[scale = 0.5]{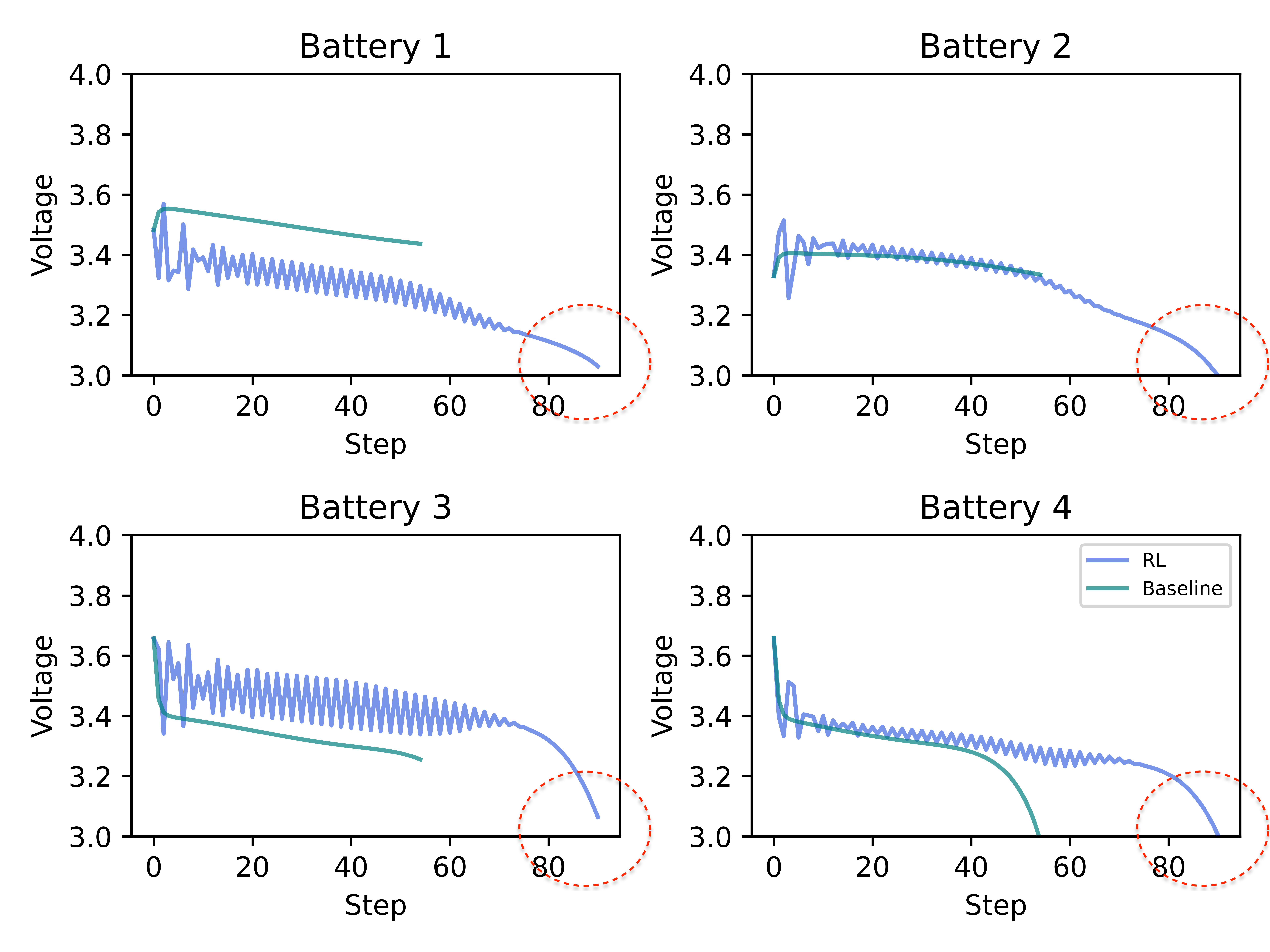}
      \caption{\textbf{Voltage trajectories} of a randomly selected test case. \textbf{A - Voltage Trajectory}: The y-axis represents the observed operational voltage of the corresponding batteries. The x-axis represents the decision-making steps.}
    \label{fig:mainre}
\end{figure*}

\begin{figure*}[h]
    \centering
    \subfigure{
    \begin{minipage}[t]{0.45\linewidth}
    \centering
    \includegraphics[scale = 0.35]{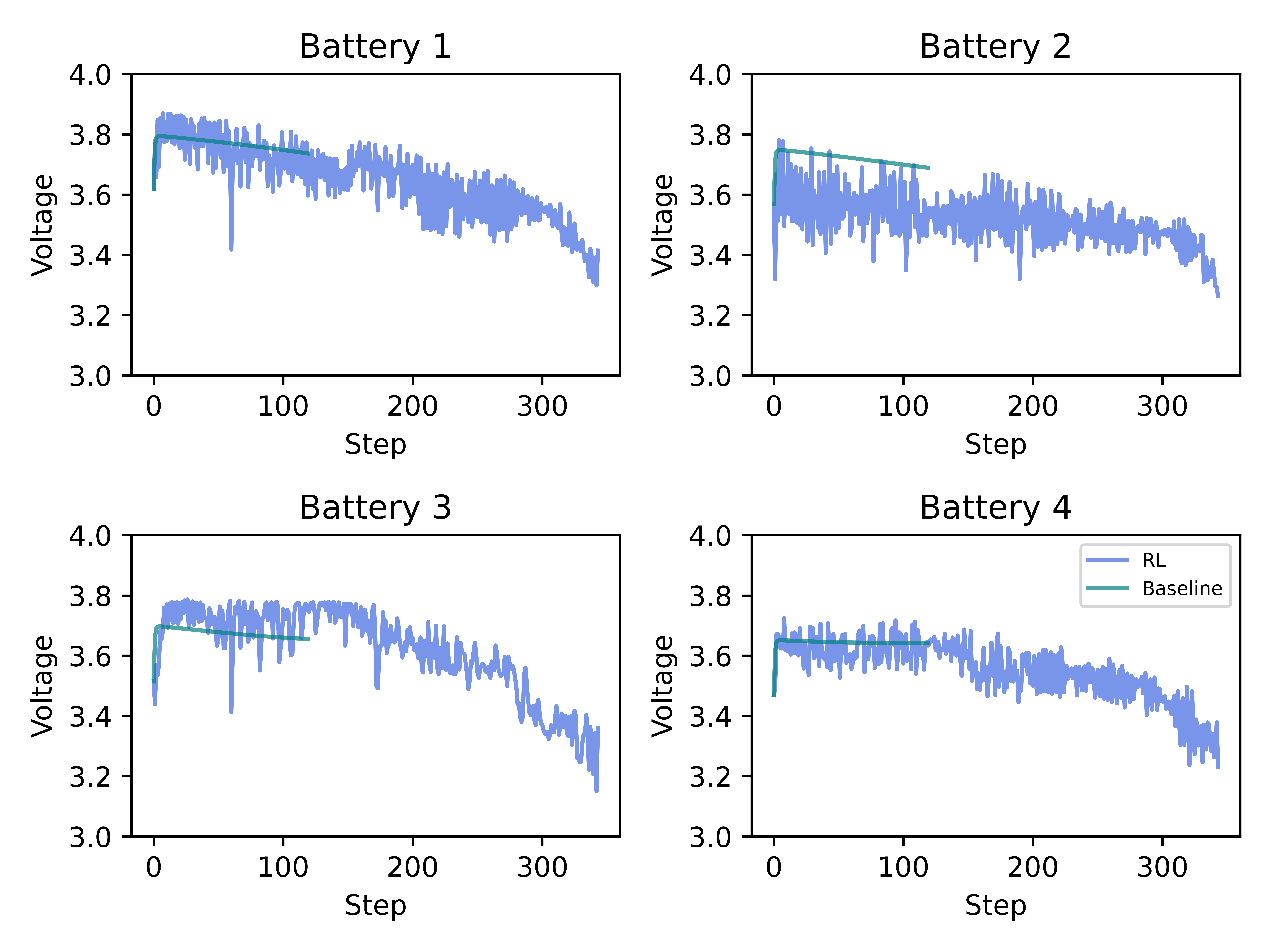}
    \end{minipage}
    }
    \subfigure{
    \begin{minipage}[t]{0.45\linewidth}
    \centering
    \includegraphics[scale = 0.35]{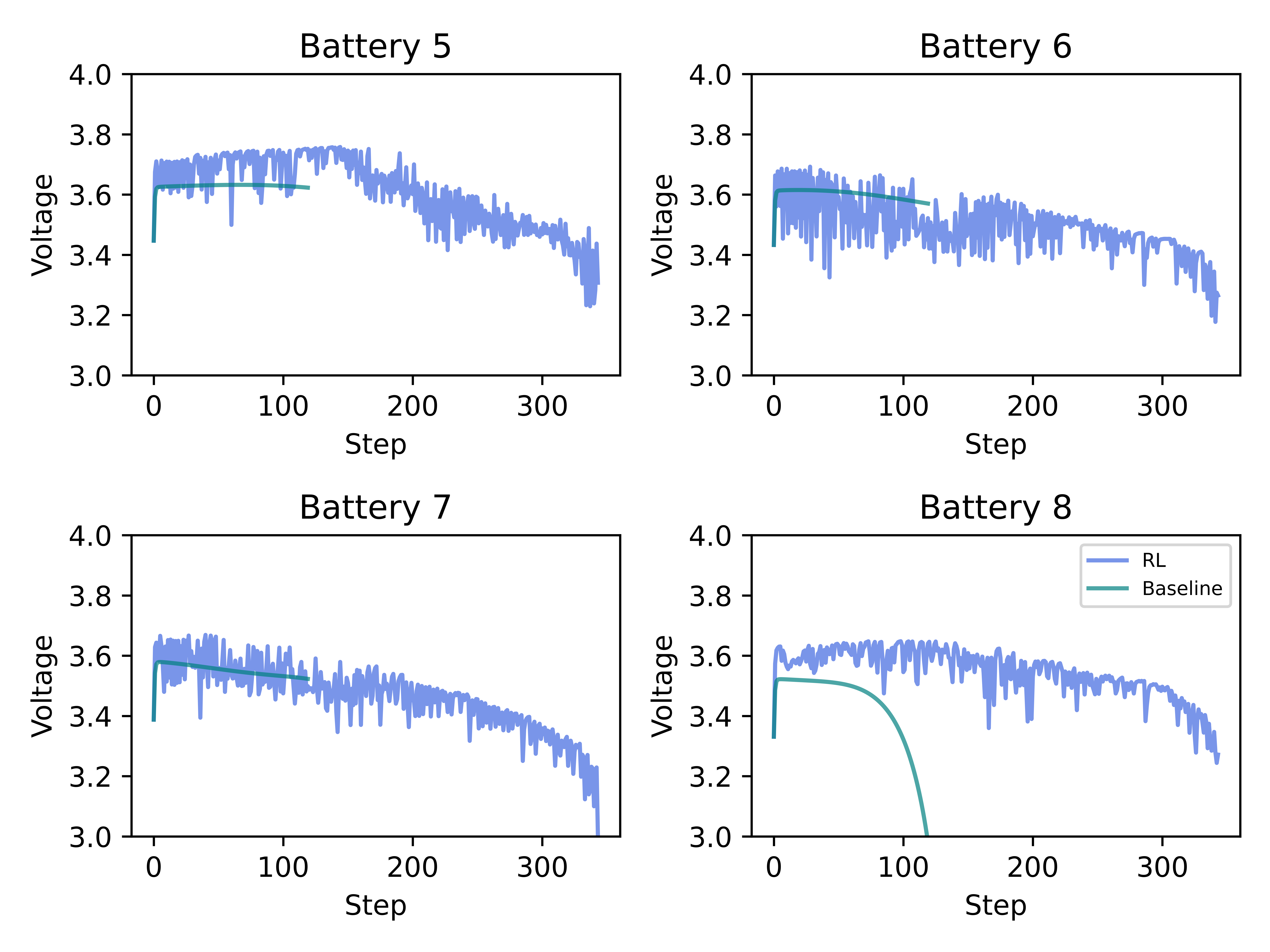}
    \end{minipage}
    }
    \caption{\textbf{Eight-battery system case result.} Figure shows a randomly selected trajectory from the 5000 random test cases. The y-axis represents the observed operational voltage of the corresponding batteries, while the x-axis represents the decision-making steps. }
    \label{fig:8}
\end{figure*}

\textit{1) Performance evaluation on a four-Li-I-battery system. }The trained strategy is tested on 5000 different random initializations with random load profiles. Compared to the baseline strategy (distributing the power equally between all the batteries), the proposed framework prolongs the working cycle by 15.2$\%$ on average. We can observe that the single batteries were controlled by the RL algorithm in such a way that they tended to reach the EoD state at approximately the same time (Fig \ref{fig:mainre}). This is an indication of a near-optimal performance. The proposed strategy also demonstrates a relatively smooth allocation profile (Fig \ref{fig:mainre}). 


\textit{2) Scalability evaluation on an eight Li-I batteries system. } Scalability is an essential requirement for power allocation approaches since different assets will have different numbers of configurations. Previous RL approaches discretize the action and state space, defining different weight combinations \cite{xiong2018reinforcement,maia2020regenerative}, which need to re-design the action space when scale up the system size. We present the proposed approach on an eight-battery system, following the same setup as for the system with four batteries and show good scalability. Since more batteries provide more flexibility, the proposed RL framework again displays superior performance compared to the baseline. The performance improvement is significantly higher compared to the four-battery case study: the lifetime can be extended by 31.9$\%$ on average over all the tested experiments in comparison to the baseline. Similar properties can be observed as in the four-battery case: the batteries can reach the EoD state nearly simultaneously (Fig \ref{fig:8}), indicating a near optimal allocation performance. The discharge curves are partly influenced by the allocation strategy. The oscillation represents the changing weights for the load allocated to each of the batteries. We observed on the performed experiments that 
the RL policy appears to prefer to frequently change the weights between the different batteries to prolong the working cycle. The investigation of this behaviour and the improvement of the interpretability will be our future work.

\begin{figure*}[h]
    \centering
    \subfigure{
    \begin{minipage}[t]{0.45\linewidth}
    \centering
    \includegraphics[scale = 0.35]{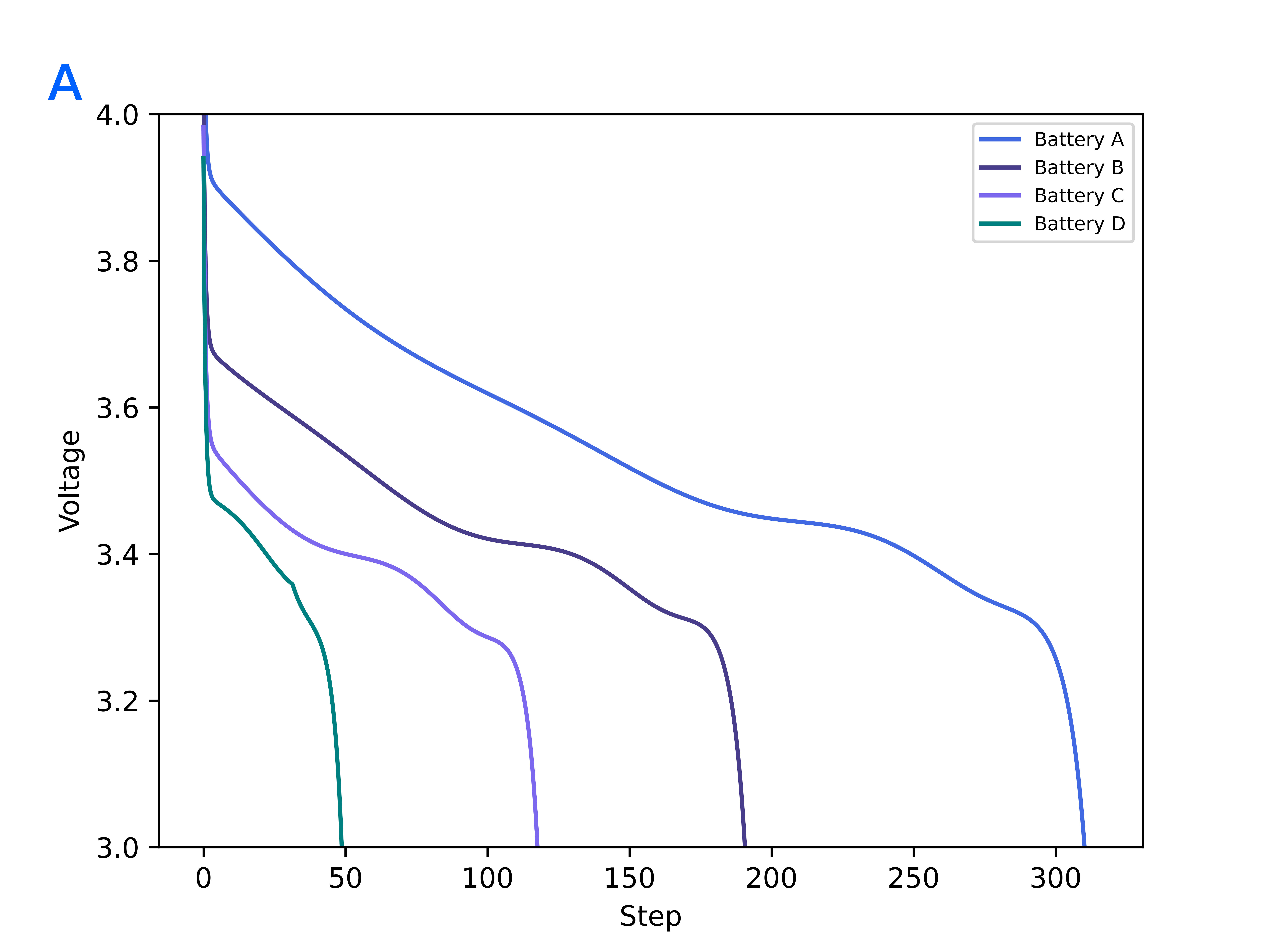}
    \end{minipage}
    }
    \subfigure{
    \begin{minipage}[t]{0.45\linewidth}
    \centering
    \includegraphics[scale = 0.335]{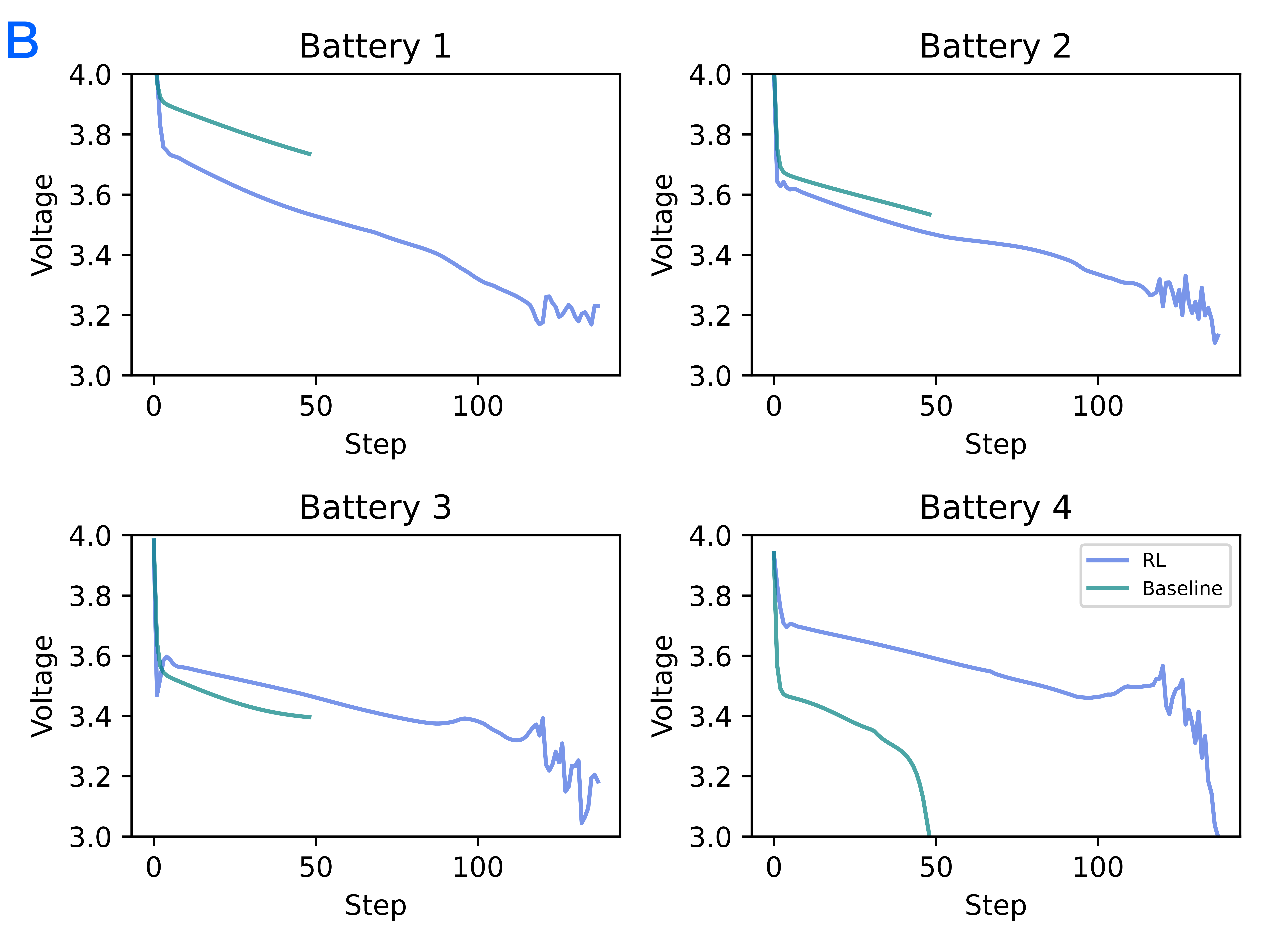}

    \end{minipage}
    }
    \caption{\textbf{Second-life battery system case result.} The y-axis represents the observed operational voltage of the corresponding batteries, while the x-axis represents the decision-making steps.}
    \label{fig:degradation}
\end{figure*}

\textit{
3) Transferablity evaluation based on a four second-life Li-I battery system} In this research, to evaluate the transferability of the proposed approach to systems with different degradation dynamics \cite{chao2022fusing}, we consider batteries in second life applications \cite{peterson2010economics,hu2020battery,fink2020potential}. Even under the same state initialization and same load profile, second-life batteries with dissimilar degradation dynamics will reach the EoD state much earlier. In Figure \ref{fig:degradation}, \textbf{A} presents the voltage trajectories of four batteries. From battery A to D, the degradation becomes more notable. Even under the same initialization and same load profile, the discharge curve changes significantly. \textbf{B} is a randomly chosen trajectory from the 5000 random test cases. The policy could significantly prolong the working cycle of the deployed
second-life battery cases. 

For this evaluation, we keep all settings similar to those from the previous two experiments. The proposed approach achieves a 151.0$\%$ improvement on average compared to the equal load distribution. 

The proposed approach demonstrates even more potential in systems with different power source dynamics and cases with high load scenarios or degraded assets.

\begin{figure}[ht]
\centering
\includegraphics[scale = 0.3]{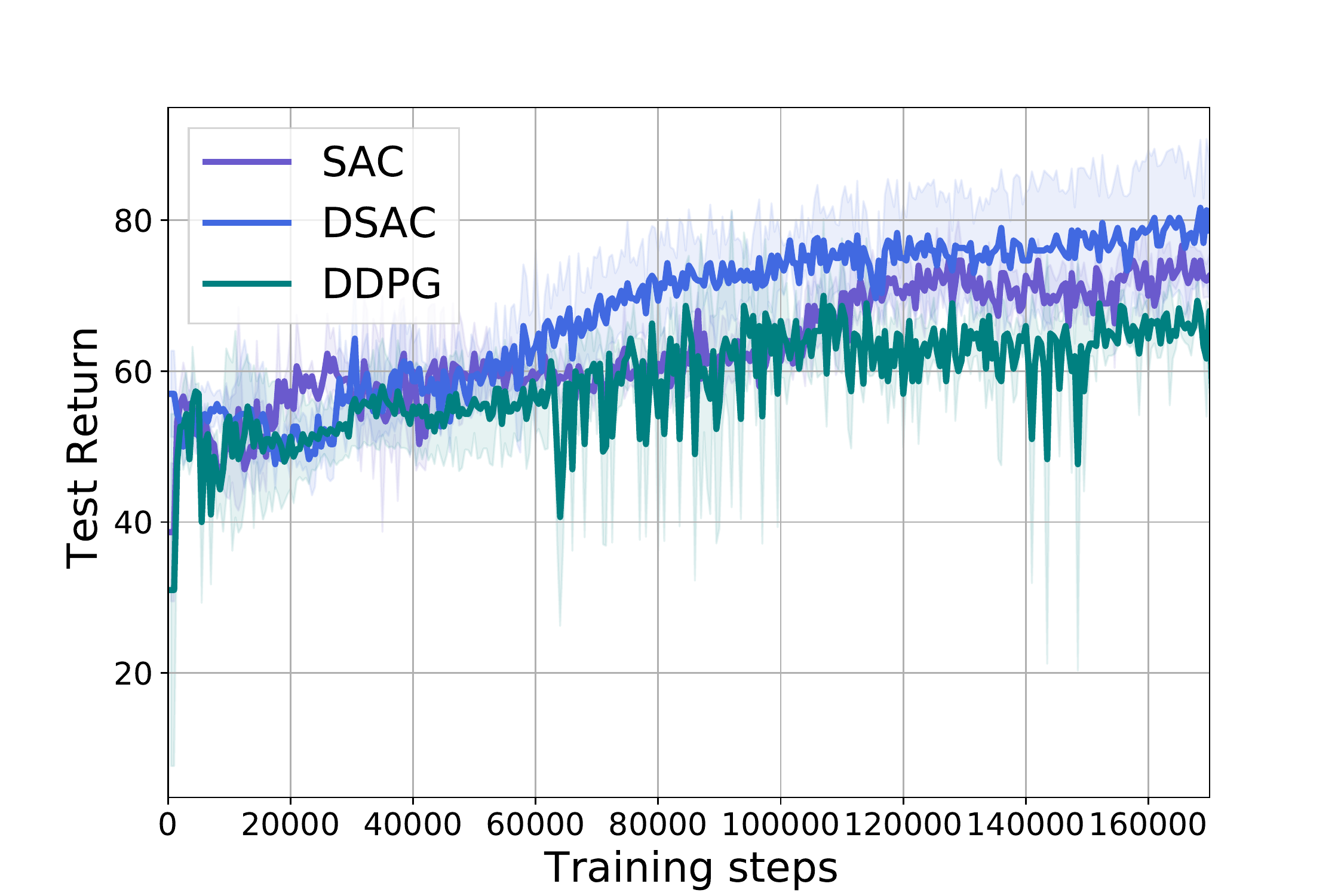}
\caption{\textbf{Average performance of the proposed Dirichlet-SAC algorithm as well the two alternative RL algorithms:  SAC and DDPG}, where the shaded areas show the standard deviation confidence intervals over three random seeds. The x-axis indicates the total time steps. And the y-axis indicates the test return.}
\label{fig:REPR}
\end{figure}

\textit{
4) Learning performance compared to SOTA} To further evaluate the performance and reproducibility of the results of the proposed Dirichlet policy, we compare it to two alternative RL algorithms: the original SAC \cite{haarnoja2018soft}, one of the state-of-the art reinforcement learning algorithms and the deep deterministic policy gradient (DDPG) \cite{lillicrap2015continuous}. We train all the agents over three different seeds on the four-battery case study. As shown in Figure \ref{fig:REPR}, we observe that the proposed Dirichlet-SAC (DSAC) shows a considerable reproducibility and also a superior performance and convergence speed compared to the original SAC and the DDPG. 

\textit{
5) Comparison to heuristic strategies} To the best of our knowledge, there are no other approaches allowing an optimization based on only on voltage-current measurements. We perform this comparison for the sake of completeness of the evaluations. We would also like to emphasize that this is not a fair comparison.  Since we would like to prolong the time to the EOD state, we define 4 heuristic strategies based on the operation voltage. We compare the average relative performance $\frac{working-cycle}{baseline-working-cycle}$ to the baseline among 5000 different random initializations with random load profiles. However, these strategies even result in worse performance compared to the baseline, see Table \ref{tb:rules}.

\begin{table}[ht]
\begin{center}
\begin{tabular}{l|c|c}
\hline
Approaches                   & Weights  & Relative Performance  \\ \hline
Rule I                  & [0.15, 0.25, 0.25, 0.35] & 0.758   \\\hline
Rule II             & [0.1, 0.2, 0.3, 0.4]  & 0.279\\ \hline
Rule III                  &  [0.1, 0.2, 0.2, 0.5]& 0.076\\\hline
Rule IV  & [0.05, 0.2, 0.35, 0.45] &0.120\\\hline
Proposed method & Learned &1.152 

\label{tb:rules}
\end{tabular}
\end{center}
\caption{Performance comparison to heuristic rules}
\end{table}

Where the weights means distributed the weighted load to the batteries with voltages from low to high respectively.

\section{Conclusion}

In this work, a novel prescriptive Dirichlet policy reinforcement learning framework is proposed for continuous allocation tasks. The proposed method overcomes the bias estimation and large variance problems in policy gradient, and can be applied to any general real-world allocation task. It is also compatible with all other continuous control reinforcement learning algorithms with stochastic policies. Besides, for a specific real-world prescriptive operation task, the power allocation task, we introduce the Dirichlet power allocation policy, which presents an effective and data-based prescriptive framework that is fully autonomous, flexible, transferable, and scalable. The developed framework has the potential to improve the efficiency and sustainability of multi-power source systems. To the best of our knowledge, it is also the first framework that enables distribution of the load in an end-to-end learning setup, without any additional inputs of, e.g., SoC estimation. For future work, we aim to apply and deploy this to more challenging and larger size real-world power allocation tasks and extend it for larger size problems to evaluate its limitations.

\section*{Acknowledgement}
The contributions of Yuan Tian and Olga Fink were funded by the Swiss National Science Foundation (SNSF) Grant no. PP00P2\_176878.

\bibliographystyle{splncs04}
\bibliography{mybib}
\end{document}